\documentclass[a4paper, 10pt, conference]{ieeeconf}      % Use this line for a4 paper

\usepackage{times}
\usepackage{epsfig}
\usepackage{graphicx}
\usepackage{amsmath}
\usepackage{amssymb}
\usepackage{textcomp}
\usepackage{multirow}
\usepackage{adjustbox}

\usepackage{multirow}
\usepackage{colortbl}
\usepackage{hhline}
\usepackage{subcaption}
\usepackage{verbatim}
\usepackage{gensymb}
\usepackage{xcolor}
\usepackage{hyperref}

\IEEEoverridecommandlockouts                              % This command is only needed if 
\title{\LARGE \bf
RGB and LiDAR fusion based 3D Semantic Segmentation \\ for Autonomous Driving
}

\author{Khaled El Madawi$^{1}$, Hazem Rashed$^{1}$, Ahmad El Sallab$^{1}$, Omar Nasr$^{2}$, Hanan Kamel$^{2}$ and Senthil Yogamani$^{3}$ \\ % 
$^{1}$Valeo R\&D Cairo, Egypt $^{2}$Cairo University $^{3}$Valeo Visions Systems, Ireland \\
         {\tt \small khaled.elmadawi@gmail.com,\{ hazem.rashed, ahmad.el-sallab, senthil.yogamani\}@valeo.com}
}

\begin{document}

\maketitle
\thispagestyle{empty}
\pagestyle{empty}

%%%%%%%%%%%%%%%%%%%%%%%%%%%%%%%%%%%%%%%%%%%%%%%%%%%%%%%%%%%%%%%%%%%%%%%%%%%%%%%%
\begin{abstract}
LiDAR has become a standard sensor for autonomous driving applications as they provide highly precise 3D point clouds. LiDAR is also robust for low-light scenarios at night-time or due to shadows where the performance of cameras is degraded. LiDAR perception is gradually becoming mature for algorithms including object detection and SLAM. However, semantic segmentation algorithm remains to be relatively less explored. Motivated by the fact that semantic segmentation is a mature algorithm on image data, we explore sensor fusion based 3D segmentation. Our main contribution is to convert the RGB image to a polar-grid mapping representation used for LiDAR and design early and mid-level fusion architectures. Additionally, we design a hybrid fusion architecture that combines both fusion algorithms. We evaluate our algorithm on KITTI dataset which provides segmentation annotation for cars, pedestrians and cyclists. We evaluate two state-of-the-art architectures namely SqueezeSeg and PointSeg and improve the mIoU score by 10\% in both cases relative to the LiDAR only baseline. 
\end{abstract}

%%%%%%%%%%%%%%%%%%%%%%%%%%%%%%%%%%%%%%%%%%%%%%%%%%%%%%%%%%%%%%%%%%%%%%%%%%%%%%%%
\section{INTRODUCTION}

Autonomous driving is a complex task where a robotic car is expected to navigate with full autonomy in a highly dynamic environment. To accomplish such task, the autonomous vehicle has to be equipped with multiple sensors and robust algorithms that perceive the surrounding environment with high accuracy in a real-time fashion. The first step for perception pipeline is to detect objects from background. Object detection alone is not sufficient for a robot to navigate, there has to be robust classification to determine the type of each object for planning the interaction, especially for complex scenarios like parking \cite{heimberger2017computer} \cite{horgan2015vision}. This is a crucial task because the reaction of an autonomous vehicle to a pedestrian that showed up suddenly after occlusion will be completely different compared to a suddenly appearing vehicle for example. Moreover, the algorithms have to estimate the location of external objects within the subsequent frames to be able to take proper action. \\

\begin{figure}[!t]
\centering
\includegraphics[width=\columnwidth]{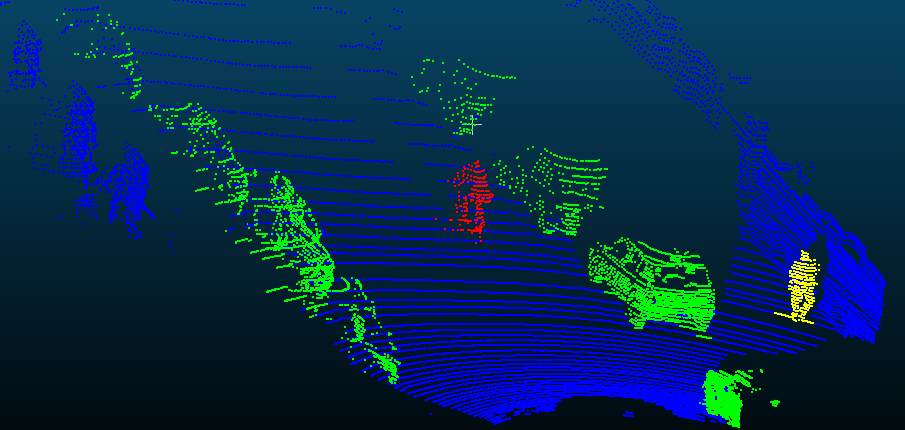}
\caption{Visualization of 3D semantic segmentation ground truth in Cartesian coordinates.}
\label{fig:CartesianPC}
\end{figure}

From this perspective, 3D semantic segmentation is a critical task for autonomous driving as it simultaneously performs 3D localization and classification of objects as visualized in Fig. \ref{fig:CartesianPC}. Point cloud segmentation has been studied in \cite{douillard2011segmentation}\cite{himmelsbach2008lidar}\cite{zermas2017fast}. Classical methods used pipelines including segmenting ground from foreground objects, clustering the objects points together and performing classification based on hand-crafted features. Such methods are prone to low performance due to accumulation of error across successive tasks. Moreover, they do not generalize well across different driving environments. %Deep learning on the other hand solves these problems effectively using a data-driven approach. 
Deep learning recently has gained large attention in several tasks including semantic segmentation due to its powerful automatic feature extraction, where it has the ability to find ambiguous relationships between different domains than sometimes cannot be interpreted by humans. In this paper, we adopt end-to-end convolutional neural network (CNN) which performs 3D semantic segmentation. Our approach utilizes both information from two complementary sensors, namely cameras and LiDAR, that are being deployed in recent commercial cars.  Camera sensor provides color while LiDAR provides depth information. Recent work in \cite{wu2018squeezeseg}\cite{wang2018pointseg} provided algorithms to understand semantics from LiDAR only. We build upon this work by fusing the color information with LiDAR data for 3D semantic segmentation task. We implement two fusion algorithms to study the impact of the addition of color information. The first approach is early-fusion in which we fuse the information as raw data before feature extraction. The second approach is mid-fusion in which we use CNN to extract features from two different modalities then perform the fusion feature level. \\

\noindent The main contributions of this paper are as follows: 
\begin{enumerate} 
    \item Construction of CNN based early and mid fusion  architectures to systematically study the effect of color information on 3D semantic segmentation.
    \item Evaluation on two state-of-the-art algorithms, namely SqueezeSeg and PointSeg, and significant improvements via fusion with camera data was achieved.
    \item Design of RGB fusion representation and hybrid fusion strategy for improving performance.
\end{enumerate} 
The paper is organized as follows. Section \ref{sec:relatedwork} reviews related work in semantic segmentation and point cloud segmentation. Section \ref{sec:methodology} discusses the various proposed architectures of our algorithm. Section \ref{sec:experiments} details the dataset used, the experimental setup and discusses the results obtained. Finally, Section \ref{sec:conc} provides concluding remarks and future work.

\section{RELATED WORK} \label{sec:relatedwork}

\subsection{Semantic Segmentation for Images}
A detailed survey for image semantic segmentation for autonomous driving is presented in \cite{siam2017deep} and efficient design techniques were discussed in \cite{briot2018analysis} \cite{siam2018rtseg}. We briefly summarize the main methods used pixel-wise classification. In \cite{grangier2009deep} patch-wise training was used for classification while in \cite{farabet2013learning} the input image was fed into Laplacian pyramid to extract hierarchical features. A deep network was used in \cite{grangier2009deep} to avoid further post processing. An  end-to-end methodology was adopted for semantic segmentation in SegNet \cite{badrinarayanan2017segnet}. In \cite{long2015fully}, the network learned heatmaps that were upsampled to generate the classification output. Multiple skip connections between the encoder and decoder were introduced to avoid losing resolution. Unlike patch-wise methods, this approach uses the full image to generate dense semantic segmentation outputs. SegNet \cite{badrinarayanan2017segnet} proposed an encoder-decoder network was deployed where the feature maps were upsampled utilizing the information of the kept indices from the corresponding encoder layer. There has been several improvements on color based segmentation but typically they don't make use of depth information.

\subsection{Semantic Segmentation For Point Clouds}
Relative to image segmentation, there is very little literature on point cloud semantic segmentation. Majority of 3D object detection literature focuses on 3D bounding box detection but it is not the best representation of objects in many scenarios. This is analogous to 2D segmentation which provides a better representation than 2D bounding box in images. The main challenge of LiDAR perception in general is the sparse nature of point cloud data. It increases the appearance complexity drastically. Thus there are many different approaches to simplify this representation including Voxels, bird-view and polar-grid map. Voxels are clustered point-cloud data and they still suffer from sparsity. Bird-view is the simplest representation which simplifies point cloud into a flat plane. However, this loses the height information and thereby important appearance features necessary for detecting objects. \\

Bird-View Lidar semantic segmentation was performed in \cite{elmadawydeep}, where the LiDAR points are projected to a grid xy-plane and a semantic classification is applied on each grid. Other approaches divided the space into voxels \cite{zhou2018voxelnet} with a predefined resolution, projected the point cloud inside these voxels, and performed voxel-level classification. However, the main problem of voxelization is the required resources in memory and processing power to represent a huge volume covered by a Lidar sensor due to considering the occupied and non-occupied voxels. Efficient implementation of moving object segmentation was explored in \cite{ravi2018real} using HD maps. On the other hand, there were other approaches that performed semantic segmentation using 3D point cloud like PointNet \cite{qi2017pointnet},  PointNet++ \cite{qi2017pointnet} which considered the point cloud as an un-ordered set of points. This approach provides invariance to arbitrary viewpoint but in autonomous driving the specific perspective is important. It also  doesn't take into consideration the structure of LiDAR scanning. Recently, Squeezeseg \cite{wu2018squeezeseg}  tackled the problem with polar-grid map which is discussed in detail in Section \ref{sec:methodology} A.  It uses the spherical representation of Lidar Point cloud which explicitly models the scanning process and it provides a relatively dense 2D plane. This has provided the opportunity to leverage image-based semantic
segmentation architectures.
% \begin{figure}[t]
% \centering
% \includegraphics[width=\columnwidth]{images/pgm_representataion/pcds_22011_09_26_0005_0000000150.png}
% \caption{Visualization of 3D semantic segmentation ground truth in Cartesian coordinates.}
% \label{fig:CartesianPC}
% \end{figure}

\begin{figure}[t]
\centering
\includegraphics[width=\columnwidth]{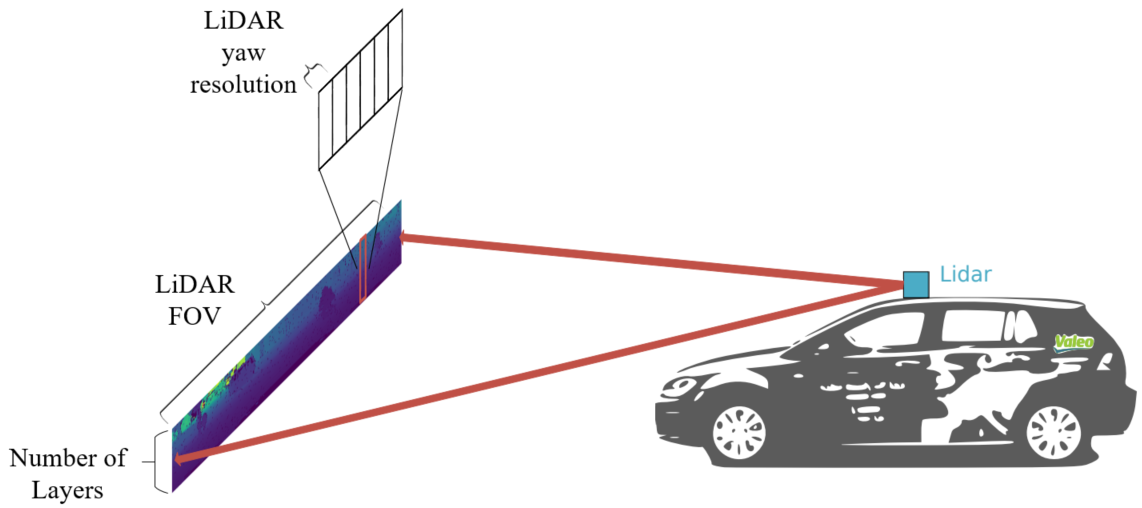}
\caption{Illustration of LiDAR Polar Grid Map representation.}
\label{fig:pgmrep}
\end{figure}

% \subsection{\textcolor{red}{LiDAR representations, PGM, Bird-eye, Voxel,....etc, advantages, disadvantages}}

\section{PROPOSED ALGORITHM} \label{sec:methodology}

\begin{figure}[!t]
\centering
\includegraphics[width=\columnwidth]{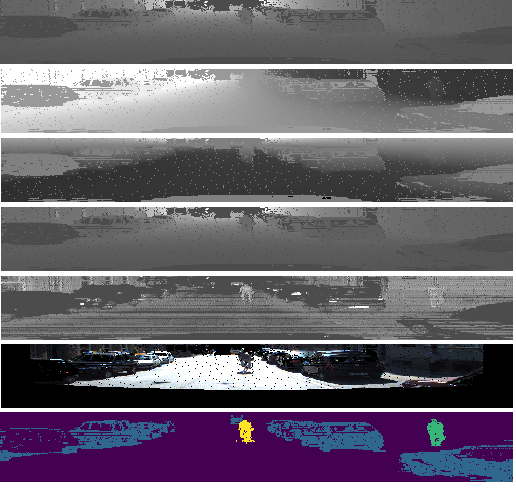}
%\vspace{5mm}
% \includegraphics[width=\columnwidth]{images/pgm_representataion/2011_09_26_0005_0000000150.png}
\caption{Input frame and ground-truth tensor. \textbf{Top to bottom:} X, Y, Z, D, I, RGB and Ground Truth.}
\label{fig:pgmrep_8channels}
\end{figure}

\begin{figure*}[!t]
\centering

\begin{subfigure}{\textwidth}
\centering
\includegraphics[width=0.8\textwidth]{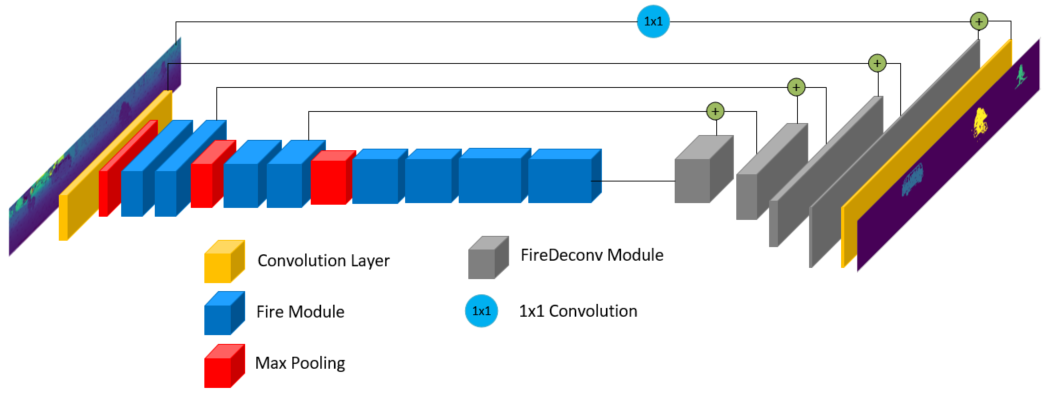}
\caption{LiDAR baseline architecture based on SqueezeSeg \cite{wu2018squeezeseg}.}
%\label{fig:unimodal}
\end{subfigure}

\begin{subfigure}{\textwidth}
\centering
\includegraphics[width=0.8\textwidth]{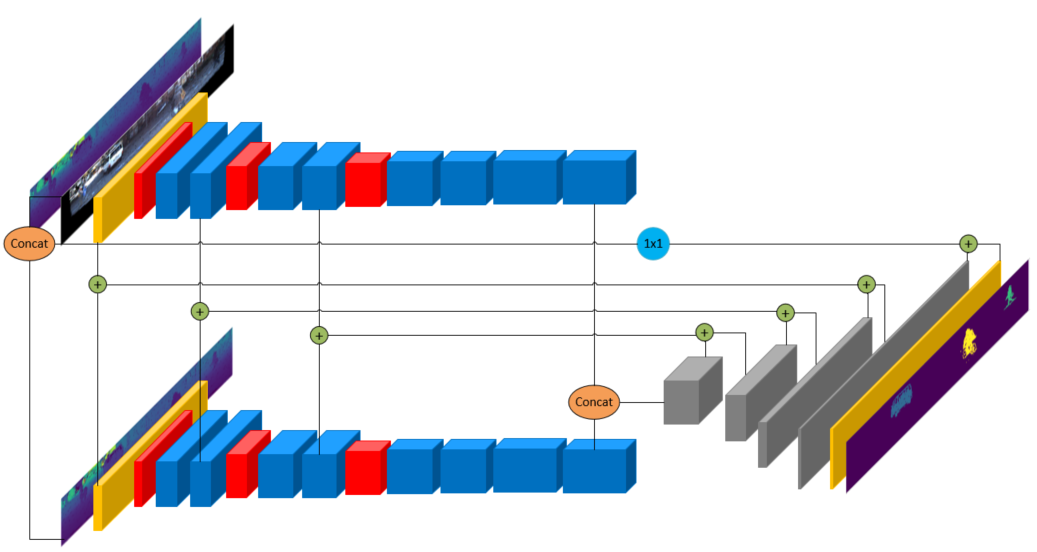}
\caption{Proposed RGB+LiDAR mid-fusion architecture}

\end{subfigure}
\caption{Semantic Segmentation network architectures used in the paper. (a) shows the baseline SqueezeSeg based unimodal baseline architecture. The architecture remains the same for early fusion except for the change in number of input planes. (b) shows the proposed mid-fusion architecture.}
\label{fig:arch}
\end{figure*}

% \begin{figure*}[!t]
% \centering
% \begin{subfigure}{\textwidth}
% \centering
%     \includegraphics[scale=0.45]{Images/arch_1} 
%     \caption{\textcolor{black}{Input RGB}} 
% \end{subfigure}%
% \qquad
% \begin{subfigure}{\textwidth}
% \centering
%     \includegraphics[scale=0.45]{Images/arch_2}
%     \caption{\textcolor{black}{Input Depth}}    
% \end{subfigure}%
% \qquad
% \begin{subfigure}{\textwidth}
% \centering
%     \includegraphics[scale=0.45]{Images/arch_3}
%     \caption{\textcolor{black}{Input RGBD}}    
% \end{subfigure}%
% \qquad
% \begin{subfigure}{\textwidth}
% \centering
%     \includegraphics[scale=0.45]{Images/arch_4}
%     \caption{\textcolor{black}{Two Stream RGB+D}}    
% \end{subfigure}%

% % \includegraphics[width=0.5\textwidth]{Images/arch_1}
% \caption{Four types of architectures constructed and tested in the paper. (a) and (b) are baselines using RGB and Depth only. (c) and (d) are depth augmented semantic segmentation architectures.}
% \label{fig:architectures}
% \end{figure*}

\subsection{Polar Grid Map representation for LiDAR and cameras} 
%One of the biggest advantage in 
LiDAR sensor can be modeled as a sorted set of points, where measurement for $n$ number of rays is captured at an instant and each ray represents a scan point in a layer. This is performed at a certain yaw angle and then the LiDAR rotates to the next yaw angle to make another measurement. This is continued to obtain a full 360$^{\circ}$ perception of the scene around the vehicle.
The point cloud comprises of a large number of points \textit{$N =$ (Number of layers) $\times$ (Number of points per layer)} where the number of points per layer \textit{$=$ (Yaw FOV) $\backslash$ (Yaw resolution of the firing angle)}.
This sparse 3D data can be reshaped into a 2D polar grid where the height is the number of layers and the width is the number of points detected per layer as shown in Fig. \ref{fig:pgmrep}.
Front-view is the most important part for driving and a subset of the full LiDAR scan for front-view is constructed for a field-of-view (FOV) of 90$^{\circ}$ with a yaw resolution of 0.175$^{\circ}$, 64 layers and 5 features (X, Y, Z, I, D) as shown in Fig. \ref{fig:CartesianPC}. X, Y, Z are the Cartesian coordinates of the Lidar sensor, I is the intensity of the scan point and D is its depth. This representation is called Polar Grid Map (PGM) and the size of this tensor is 64x512x5.

Camera 2D image is obtained by projection of the light rays onto the image plane. Typically, cameras may be have a different FOV compared to LiDAR. For example, LiDAR cannot view the near-field of the vehicle. The ideal-pinhole projection model of a camera is also broken by modern lenses which is very pronounced in the case of fisheye lenses. LiDAR is typically projected onto the camera image plane \cite{kumar2018monocular}. However this leads to a sparse representation of LiDAR which might lead to sub-optimal models. In this paper, we explore an alternate approach of re-projecting the pixels onto the LiDAR polar-grid map plane which is dense. We project the LiDAR point cloud on the RGB image using relative calibration information. This establishes a mapping between LiDAR scan points and RGB pixels.
% We obtain an RGB value for each scan point based on corresponding angular rays calculated from relative calibration. 
By using this mapping, we augment three additional features (RGB) to the existing point cloud feature tensor (XYZID), creating a tensor of size 64x512x8 as shown in Fig. \ref{fig:pgmrep_8channels} and the corresponding ground truth in a tensor of size 64x512x1 is shown at the bottom of Fig. \ref{fig:pgmrep_8channels}. This representation is also scalable to map multiple cameras around the car which can cover the full 360$^{\circ}$. However, it has the disadvantage of not utilizing color pixels which do not have a mapping to a LiDAR point.

% \begin{figure}[!t]
% \centering
% \includegraphics[width=\columnwidth]{images/pgm_representataion/2011_09_26_0005_0000000150.png}
% \caption{frame ground truth tensor }
% \label{fig:pgmrep_gt}
% \end{figure}

\subsection{LiDAR baseline architecture} 
Our baseline architecture is based on SqueezeSeg \cite{wu2018squeezeseg} which is a lightweight architecture that performs 3D semantic segmentation using LiDAR point cloud only. The network architecture is illustrated in Fig. \ref{fig:arch} (a).
It is built upon SqueezeNet \cite{iandola2016squeezenet} where the encoder is used until fire9 layer to extract features from the input point cloud. Fire units include 1x1 conv layer which squeezes the input tensor to one quarter of its original depth. Two parallel convolutional layers are followed by a concatenation layer to expand the squeezed tensors. The output is passed to fireDeconv modules to do the upsampling where a deconvolutional layer is inserted between the squeezing and expanding layers. Skip connections are utilized to join the deep and shallow layers in order to maintain the high resolution feature maps and avoid accuracy loss. Finally, the output probability is generated through a softmax layer after final convolutional layer. The input to this architecture is a 5-channel tensor which includes three layers for X,Y,Z that describe the spatial location of each point in the point cloud. The fourth layer is a depth map which shows the polar distance between the LiDAR sensor and the target point. Finally, the fifth layer encodes reflectance intensity of LiDAR beams. We refer to this input as XYZDI. We also implemented another baseline architecture using PointSeg \cite{wang2018pointseg} which improves upon SqueezeSeg.

\subsection{Early-fusion architecture} 
In this architecture, we aim to fuse the input data as raw values which will be processed for joint feature extraction by the CNN. The same methodology described in the baseline network architecture is used, however the input tensor in this case consists of 8 channels which are the original XYZDI in addition to 3 RGB layers. The advantage of this architecture is that the network has the potential to learn relationships among data and combine them effectively. However, this architecture cannot directly leverage pre-training on large unimodal datasets like ImageNet \cite{deng2009imagenet} due to different number of input layers. We refer to this architecture as XYZDIRGB and we obtain improved results over the baseline architecture with negligible increase in computational complexity.

\subsection{Mid-Fusion architecture} 
We construct a mid-fusion architecture where the fusion happens at the CNN encoder feature level as illustrated in Fig. \ref{fig:arch} (b).
In this case, two separate encoders are constructed for each input modality. Feature extraction is performed on each input separately, then the processed feature maps are fused together using the concatenation layer. This model is computationally more complex than early fusion model as the the number of encoder parameters are doubled. But when it is viewed from the system level, the separate encoders can be leveraged for other tasks in the respective modalities. This model typically provides better performance compared to early fusion \cite{rashed2019optical} \cite{rashed2019motion}. We refer to this architecture as XYZDI + DIRGB. It was experimentally found that this architecture was not able to effectively fuse the modalities and there was negligible increase in accuracy. We constructed a hybrid of early-fusion and mid-fusion network where we concatenate the RGB channel to LiDAR depth and intensity channels. We obtain significant improvements over the baseline using this approach.
\section{EXPERIMENTS} \label{sec:experiments}
In this section we provide details about the dataset used and our experimental setup.

\subsection{Datasets} 
We make use of the KITTI raw \cite{Geiger2013IJRR} dataset which contains 12,919 frames, of which 10128 were chosen as training data, and  2,791  frames were used as validation data. We choose this dataset for multiple reasons. Firstly, it is based on autonomous driving scenes which is the main focus of our work. Additionally, it provides 3D bounding box annotation for multiple classes. Following \cite{wu2018squeezeseg}, we divide our classes into three groups, i.e. "Car", "Pedestrian", "Cyclist" and "Background". The "Car" class includes cars, vans and trucks. We focus on these classes because they have the main collision risk for an autonomous vehicle. The points inside each 3D bounding box are labeled with the class provided by the dataset which can be used as annotation for 3D semantic segmentation. We make use of the data split provided by \cite{wu2018squeezeseg} so that it can be compared effectively. 

\subsection{Experimental Setup}
% as mentioned before our target is to benchmark fussed lidar point cloud with camera with it's different combinations(mid fusion, and early fusion), and lidar with no fusion. our We data representation was 
We make use of PGM data representation with horizontal front FOV of 90$^{\circ}$, creating a 3D tensor of 64$\times$512$\times nc$ where $nc$ denotes the number of input channels depending on the experiment at hand. In the baseline experiments, $nc$ is 5 encoding LiDAR data only, and in Early-Fusion $nc$ is 8 encoding RGB layers concatenated to LiDAR channels. In Mid-Fusion $nc$ is 5 in the DIRGB branch and 5 in the LiDAR branch. The output is a 64x512x1 tensor representing classification per polar gird. We used data augmentation by randomly flipping the frames in the y-axis. In all experiments, we set learning rate to 0.01 and the optimizer momentum was set to 0.9. Class-wise Intersection over Union (IoU) is used as the performance metric, and an average IoU is computed over all the classes. Our model is implemented using TensorFlow library. We ran our training and inference on a 1080-ti GPU.

\begin{table}[]
\centering
\caption{Quantitative evaluation on KITTI Raw dataset using SqueezeSeg and PointSeg architectures. }
\label{tab:results}
\resizebox{0.5\textwidth}{!}{
\begin{tabular}{|l||l|l|l|l||l|}
\hline
\multicolumn{1}{|c||}{Network Type} & \multicolumn{1}{c|}{Car}       & \multicolumn{1}{c|}{Pedestrian} & \multicolumn{1}{c|}{Cyclist}   & \multicolumn{1}{c||}{mIoU}      & \multicolumn{1}{c|}{\begin{tabular}[c]{@{}c@{}}Runtime \\ (ms)\end{tabular}} \\ \hline
\multicolumn{6}{|c|}{SqueezeSeg based architectures}                                                                                                                                                      \\ \hline
XYZDI                              & 62.2                           & 16.9                            & 21.9                           & 33.7                           & 8                                 \\ \hline
XYZDIRGB                           & \textbf{65.7} & 20.2                            & 24.2                           & 36.7                           & 8                                 \\ \hline
XYZDI + DIRGB                        & 65.1                           & \textbf{22.7}  & \textbf{24.4} & \textbf{37.4} & 11                                \\ \hline
\multicolumn{6}{|c|}{PointSeg based architectures}                                                                                                                                                        \\ \hline
XYZDI                              & 67                             & 18.4                            & 19.12                          & 34.8                           & 9                                 \\ \hline
XYZDIRGB                           & \textbf{68.5} & 16.2                            & \textbf{28.8} & \textbf{37.8} & 9                                 \\ \hline
XYZDI + DIRGB                        & 67.8                           & \textbf{18.6}  & 26.3                           & 37.6                           & 12                                \\ \hline
\end{tabular}
}
\end{table}

\subsection{Results} \label{sec:results}

Table \ref{tab:results} (top) shows quantitative evaluation for our approach using SqueezeSeg architecture \cite{wu2018squeezeseg}. XYZDI results are obtained by training the publicly available network \cite{wu2018squeezeseg} using KITTI-raw dataset without fusion. These results serve as a baseline for our experiments for comparative purpose. Results of XYZDIRGB show enhanced performance over the baseline with an absolute increase of 3\% in mean IoU. XYZDI + DIRGB refers to our proposed algorithm which provides the best performance with an absolute increase of 3.7\% in mean IoU. Relative increase in mean IoU is around 10\%. Results using PointSeg \cite{wang2018pointseg} architecture are reported in Table \ref{tab:results} (bottom). Early and Mid-Fusion significantly outperform results using LiDAR data only. However, results of early fusion outperformed results of mid fusion. This result is not consistent with the previous experiments with SqueezeSeg and our prior experience on fusion architectures \cite{rashed2019optical}. After careful re-experiments to cross verify the result, we hypothesize that this could be due to the atypical enlargement layer in PointSeg network which is concatenated with the regular convolutional layer features. \\

% This was un-expected as it is not consistent with previous experieand thus careful re-experiments were performed to cross check the result. 
% We attempted further analysis but we could not interpret why this result was not consistent with previous experiments with SqueezeSeg.

It is observed that no-fusion approach had difficulties in inferring classes with small volume. We believe there are three reasons for this. The first one is the unbalanced dataset especially with the proposed split provided by \cite{wu2018squeezeseg} where only 35\% of Pedestrian class is used for training, and 65\% for testing. In Cyclist class, 63\% were used for training, and 37\% for testing. On the other hand, the Car class is divided into 78\% for training, and 22\% for testing. The second reason is the unbalanced classes, where the Car class represents 96\% of the annotated dataset, while Cyclist class represents only 1.4\% of the annotated data, and pedestrian class represents also around 1.6\% of the annotated data. The third reason is the small volume of the instances from the two classes compared to the Car class which minimizes the strong features specific to those classes. We believe these reasons played an important role in the detection problem. However both early or mid-fusion experiments provide enhanced performance over results with LiDAR only. In Pedestrian class we obtained 3.3\% and 5.8\% respectively in early and mid-fusion. In Cyclist class the mIoU was improved by 2.3\% and 2.5\% respectively for both fusion approaches. Using PointSeg architecture, we obtained 3\% and 2.8\% improvement. \\

\begin{figure*}[htpb]
\centering

\includegraphics[width=.245\textwidth, height =3.8cm]{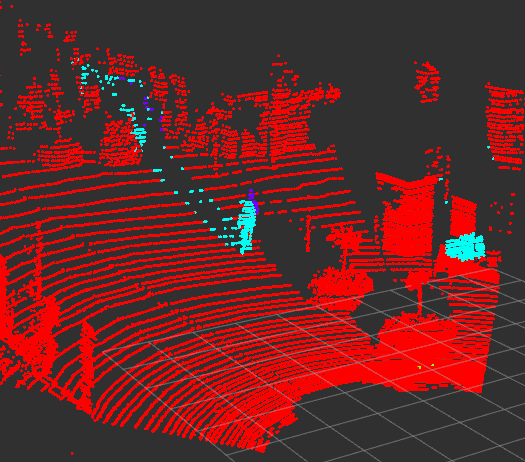}
\includegraphics[width=.245\textwidth, height =3.8cm]{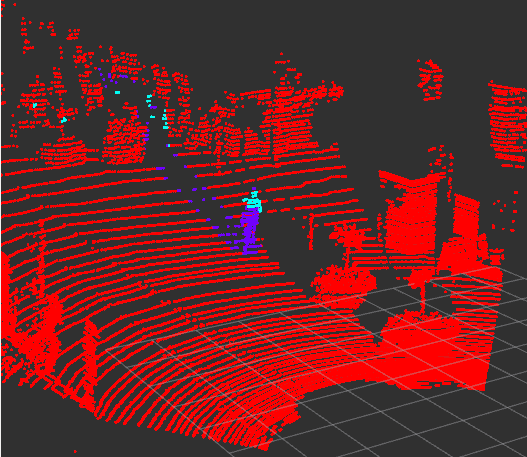}
\includegraphics[width=.245\textwidth, height =3.8cm]{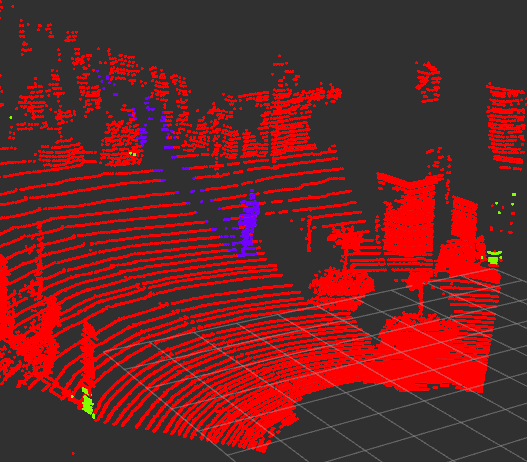}
\includegraphics[width=.245\textwidth, height =3.8cm]{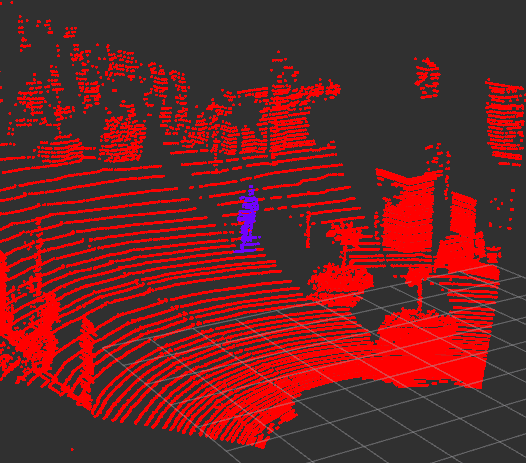}
\vspace{-3mm}

\includegraphics[width=.245\textwidth, height =3.65cm]{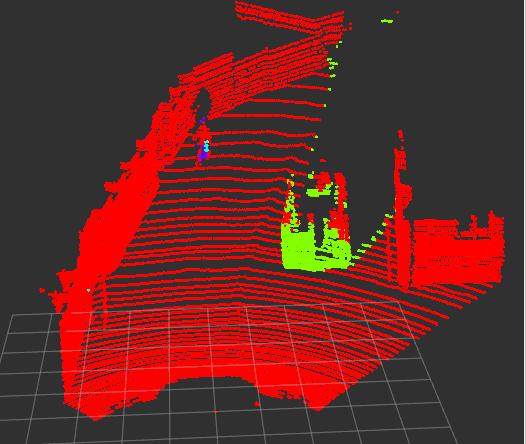}
\includegraphics[width=.245\textwidth, height =3.65cm]{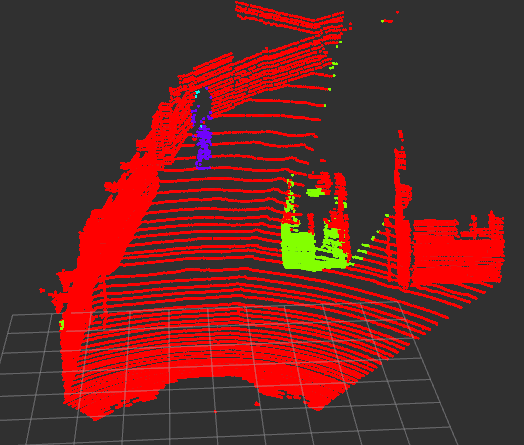}
\includegraphics[width=.245\textwidth, height =3.65cm]{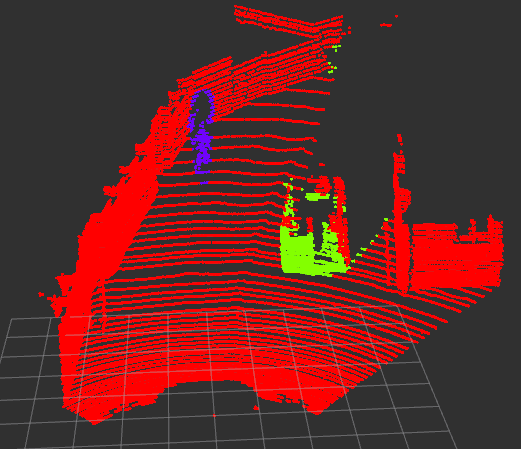}
\includegraphics[width=.245\textwidth, height =3.65cm]{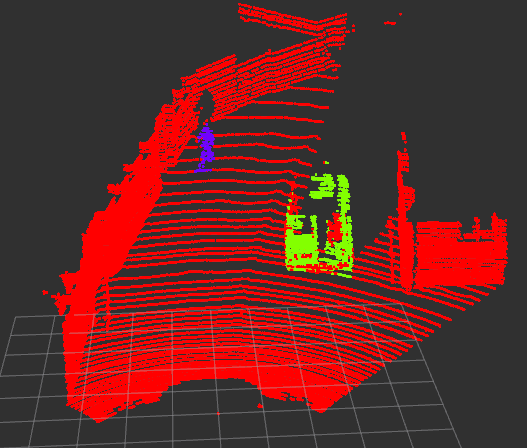}
\vspace{-3mm}

\includegraphics[width=.245\textwidth, height =5.15cm]{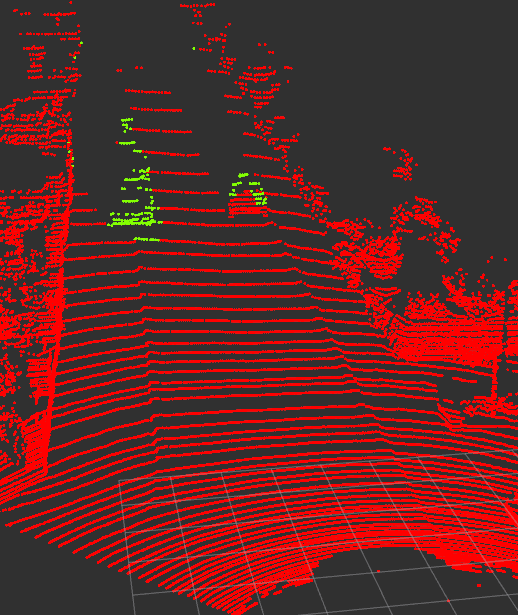}
\includegraphics[width=.245\textwidth, height =5.15cm]{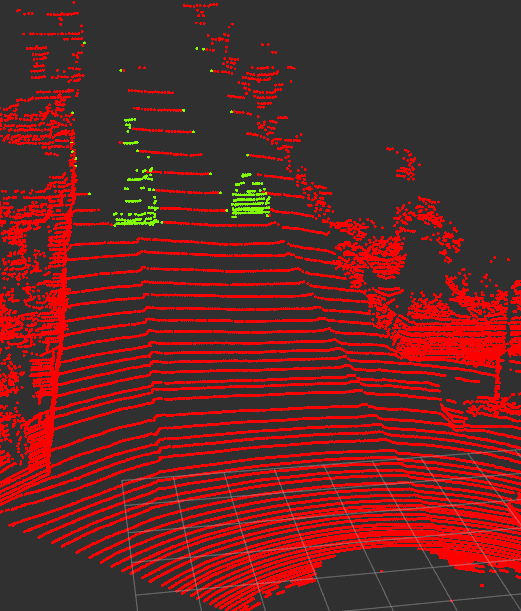}
\includegraphics[width=.245\textwidth, height =5.15cm]{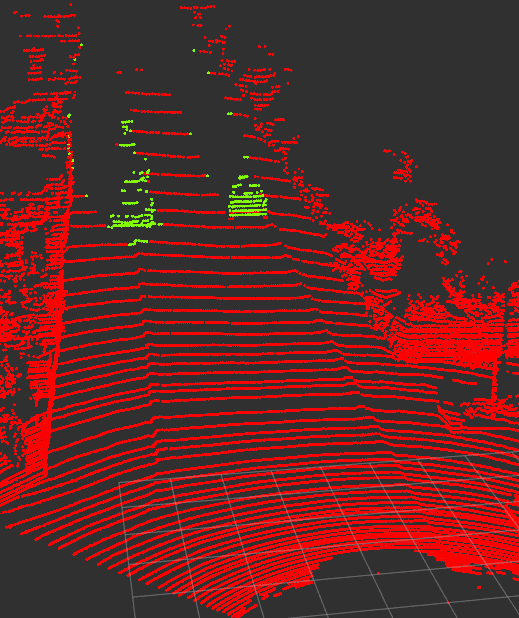}
\includegraphics[width=.245\textwidth, height =5.15cm]{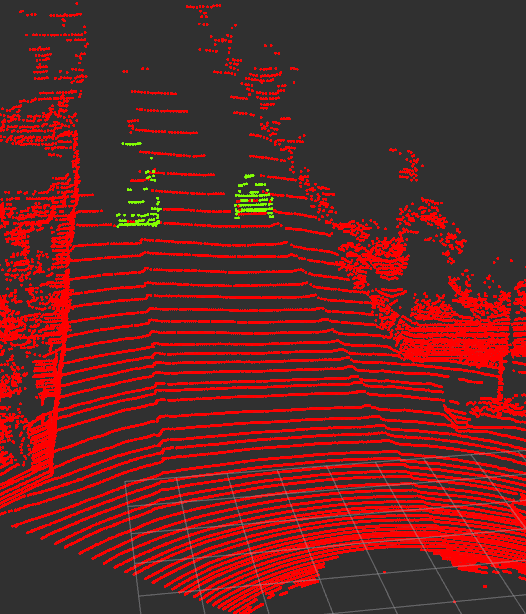}

\caption{Qualitative comparison of 3D semantic segmentation outputs using our approach on SqueezeSeg architecture where red color represents None, green color represents Cars, violet represents Cyclists, and light-blue color represents Pedestrians. \textbf{Left:} No-Fusion output. \textbf{Middle-Left:} Early-Fusion output. \textbf{Middle-Right:} Mid-Fusion output. \textbf{Right:} Ground Truth.}
    \label{fig:squeezeSeg_qualitative}
\end{figure*}

Fig. \ref{fig:squeezeSeg_qualitative} shows qualitative comparison between the results obtained using SqueezeSeg architecture. It is shown that our approach improved the detection of cars, pedestrians and cyclists using early and mid-fusion which are illustrated in the second and third columns. In the first and second rows in Fig. \ref{fig:squeezeSeg_qualitative}, the no-fusion approach classified the cyclist as a pedestrian, where early-fusion provided better accuracy with incorrect classification for the head part. Mid-fusion classified the cyclist correctly, however we notice some false positives at the edges of the cyclist, which we believe to be due to the effect of smoothing effect of convolutional filters. In the third row, Early fusion and mid fusion achieved the best classification of the car. \\

Due to the light-weight architecture, the performance of our algorithm is real-time taking around 10 ms per scan. Early fusion nearly takes the same execution time taken by the no-fusion approach, while the Mid-fusion costed 3 ms more in both architectures. Runtime details are tabulated in last column of Table \ref{tab:results}.

% \begin{table}[ht!]
% \centering
% \caption{Quantitative evaluation on KITTI Raw dataset using SqueezeSeg fusion. }
% \begin{tabular}{|l|l|l|l|l|}
% \hline
% Network Type & Car & Pedestrian & Cyclist & mIoU\\ \hline
% XYZDI & 62.2	& 16.9 & 21.9	& 33.7\\ \hline
% XYZDIRGB & \textbf{65.7}	& 20.2	& 24.2	& 36.7\\ \hline
% XYZDI + RGB & 65.1 &	\textbf{22.7} & \textbf{24.4} & \textbf{37.4} \\ \hline
% \end{tabular}
% \label{table:squeezeSeg_results}
% \end{table}

% \begin{table}[ht!]
% \centering
% \caption{performance benchmarking }
% \begin{tabular}{|l|l|}
% \hline
% Network Type & Avg runtime (ms)\\ \hline
% squeze Seg No fusion & 8\\	 \hline
% squeze Seg Early fusion & 8\\ \hline
% squeze Seg Mid fusion & 11 \\ \hline
% Point Seg No fusion & 9 \\ \hline
% Point Seg Early fusion & 9 \\ \hline
% Point Seg Mid fusion & 12 \\ \hline

% \end{tabular}
% \label{table:benchmark}
% \end{table}

% \begin{table}[ht!]
% \centering
% \caption{Quantitative evaluation on KITTI Raw dataset using PointSeg baseline architecture.}
% \begin{tabular}{|l|l|l|l|l|}
% \hline
% Network Type & Car & Pedestrian & Cyclist & mIoU \\ \hline
% XYZDI & 67 & 18.4 & 19.12 &	34.8 \\ \hline
% XYZDIRGB & \textbf{68.5}	& 16.2	& \textbf{28.8}	& \textbf{37.8}\\ \hline
% XYZDI + RGB & 67.8 &	\textbf{18.6} & 26.3 & 37.6 \\ \hline
% \end{tabular}
% \label{table:pointSeg_results}
% \end{table}

\section{CONCLUSIONS} \label{sec:conc}
In this paper, we explored the problem of leveraging color information in addition to LiDAR point clouds for 3D semantic segmentation task for autonomous driving. We remapped RGB images to LiDAR polar-grid mapping representation and constructed early and mid-level fusion architectures. We provided experimental results on KITTI dataset and improved two state-of-the-art algorithms SqueezeSeg and PointSeg by 10\% in both cases. In future work, we plan to explore more sophisticated fusion architectures using network architecture search techniques and utilize all the available color pixels.



% \section*{ACKNOWLEDGMENT}

%%%%%%%%%%%%%%%%%%%%%%%%%%%%%%%%%%%%%%%%%%%%%%%%%%%%%%%%%%%%%%%%%%%%%%%%%%%%%%%%

%\bibliographystyle{unsrt}
\bibliographystyle{ieee}
\bibliography{references/egbib}

\onecolumn
\section*{SUPPLEMENTARY MATERIAL} \label{sec:supplement}
% COMMENTED TO REDUCE TO 6 PAGES

\begin{figure*}[htpb]
\centering

\includegraphics[width=.245\textwidth, height =3.8cm]{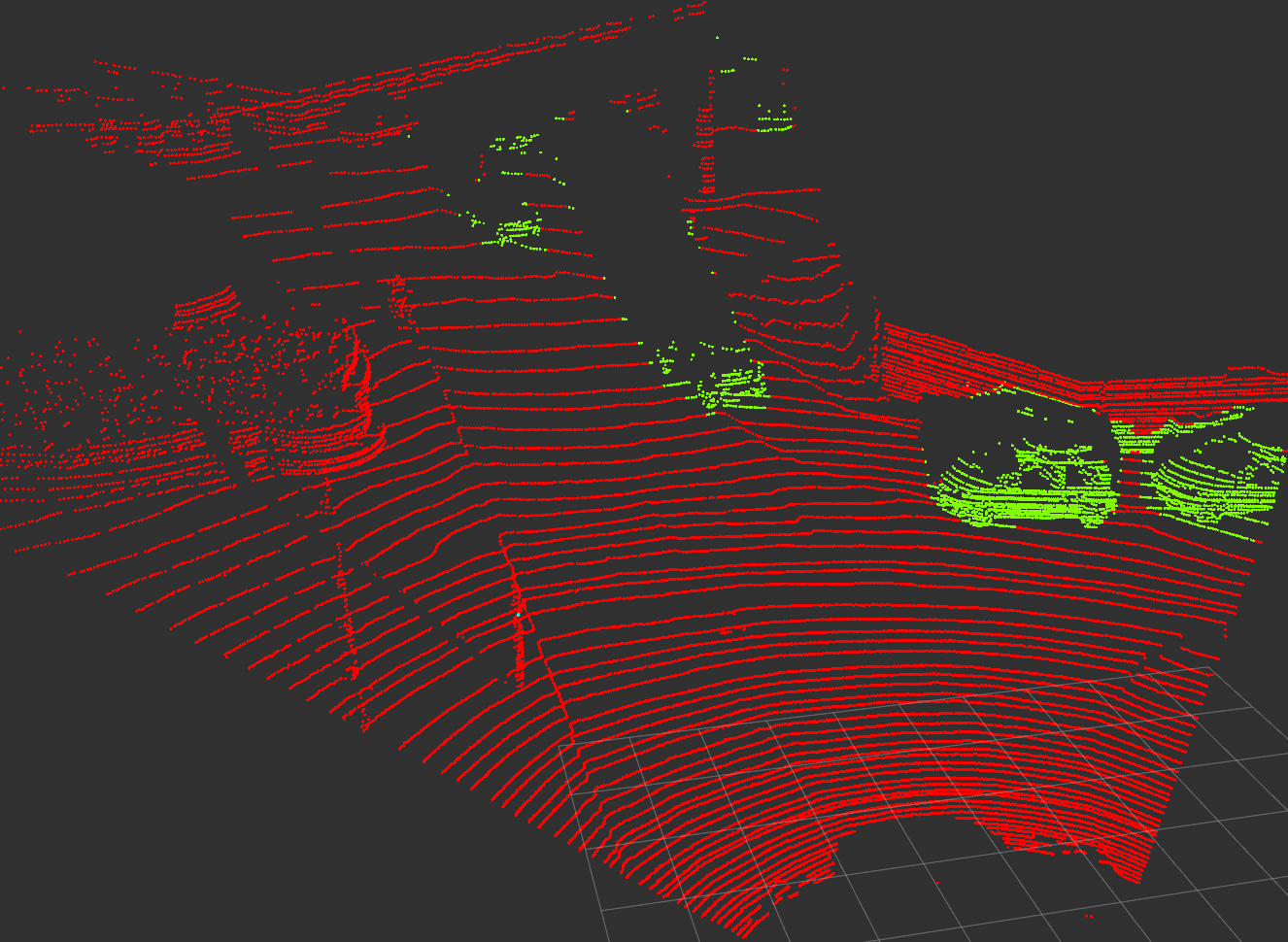}
\includegraphics[width=.245\textwidth, height =3.8cm]{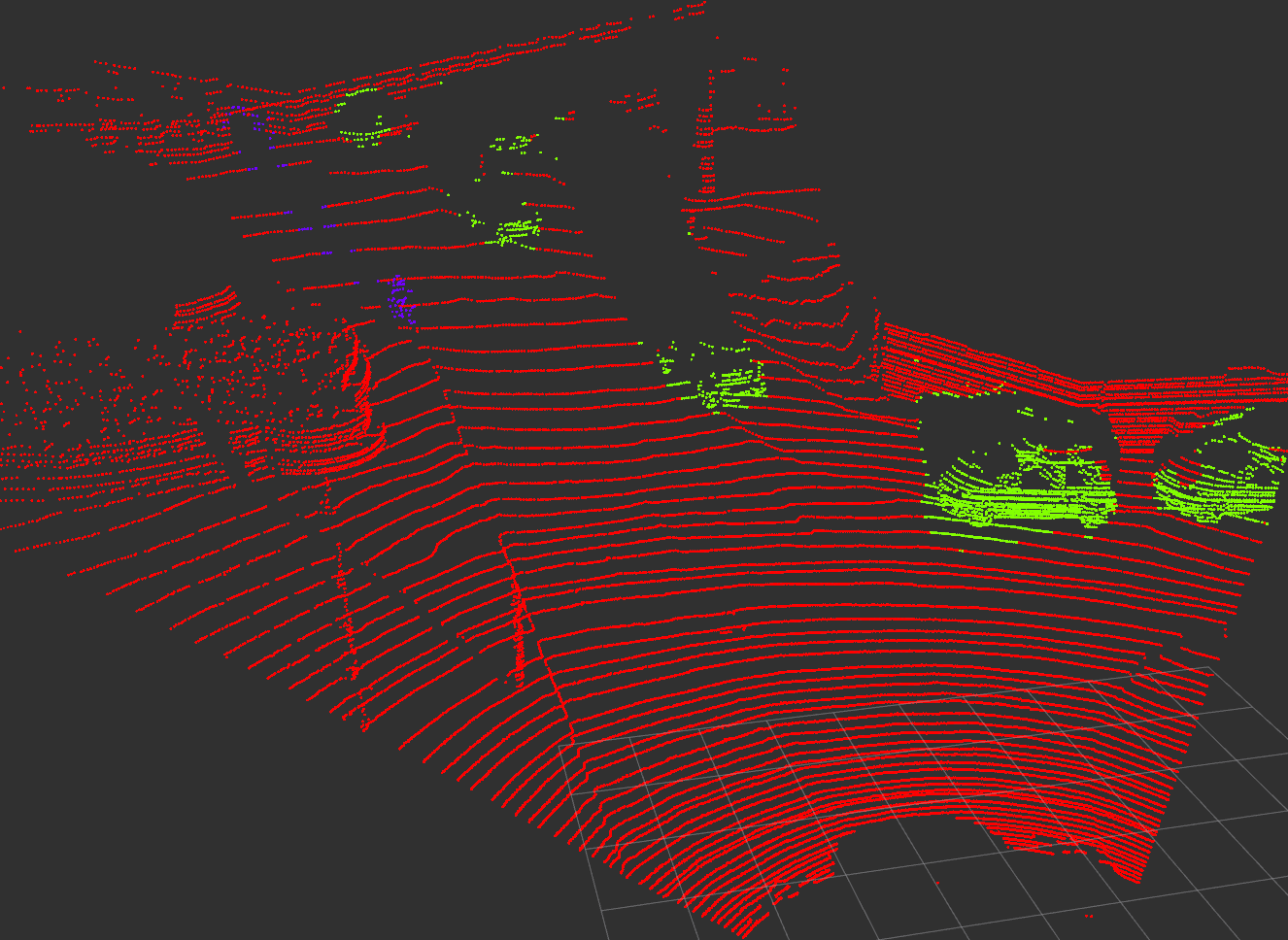}
\includegraphics[width=.245\textwidth, height =3.8cm]{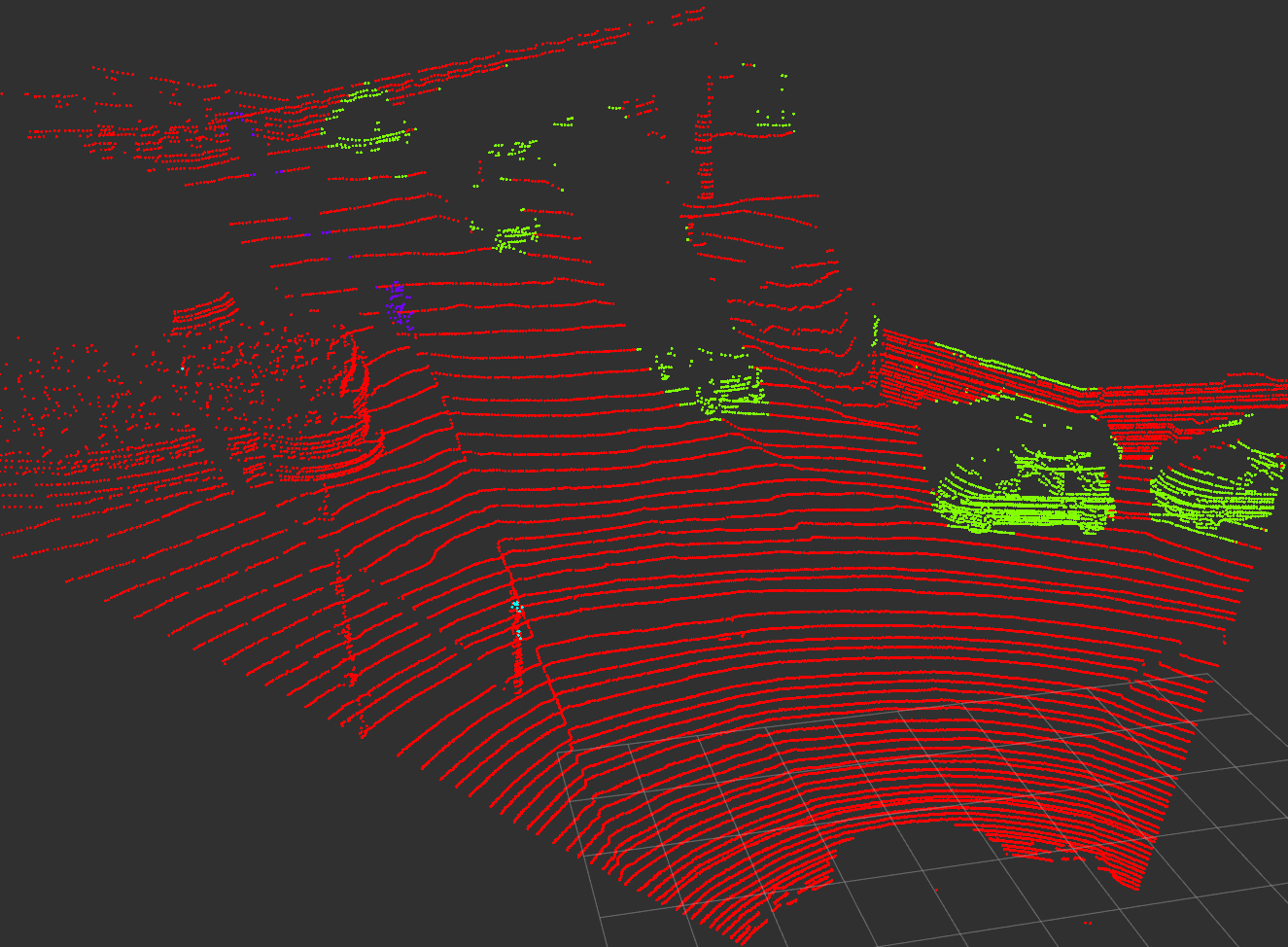}
\includegraphics[width=.245\textwidth, height =3.8cm]{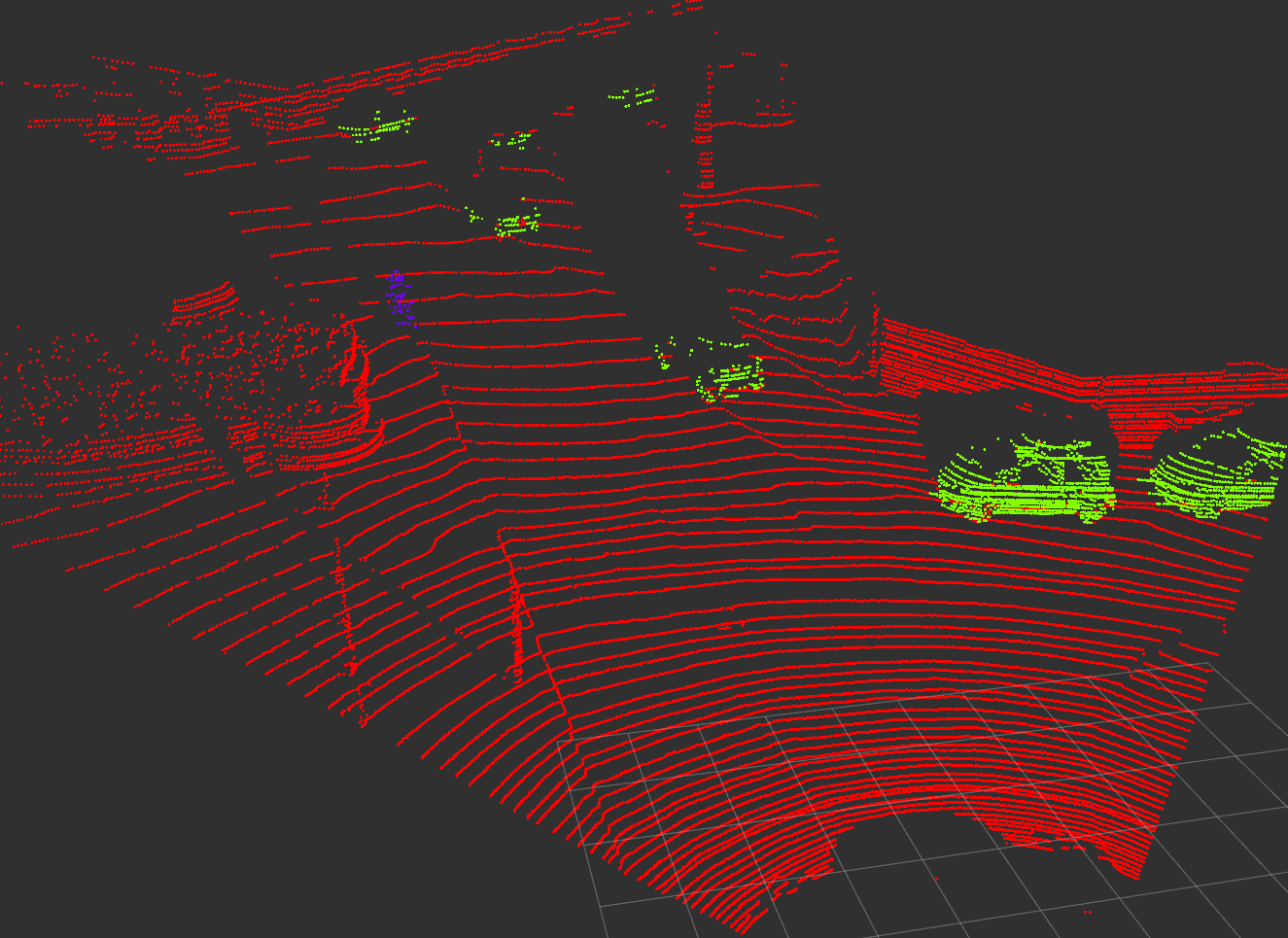}
\vspace{-3mm}

\includegraphics[width=.245\textwidth, height =3.65cm]{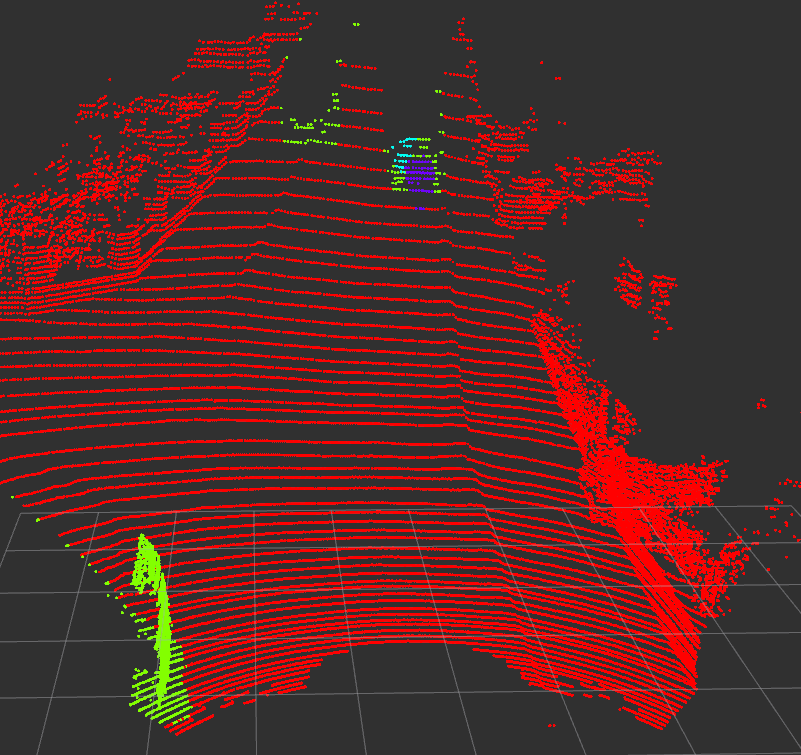}
\includegraphics[width=.245\textwidth, height =3.65cm]{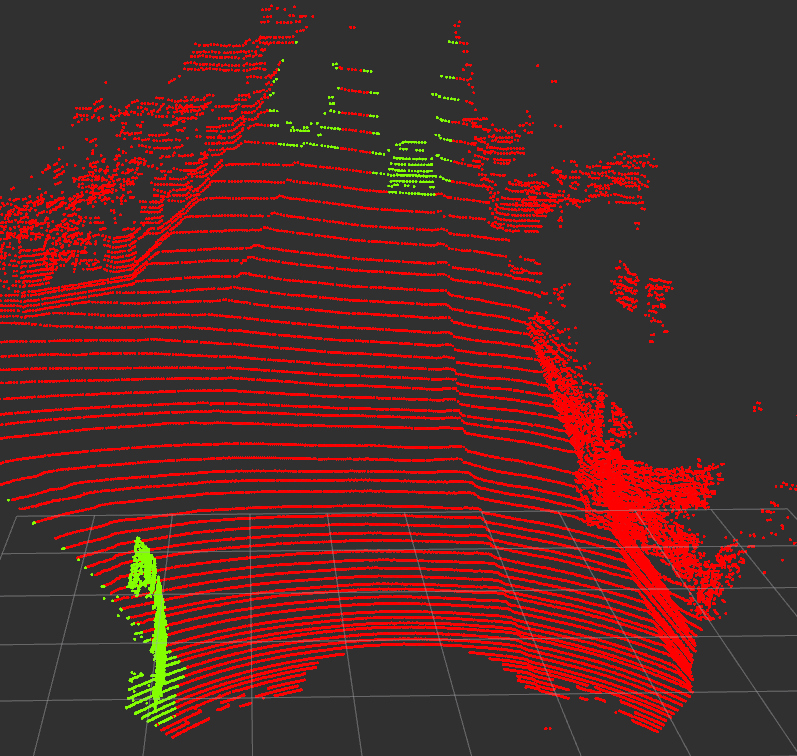}
\includegraphics[width=.245\textwidth, height =3.65cm]{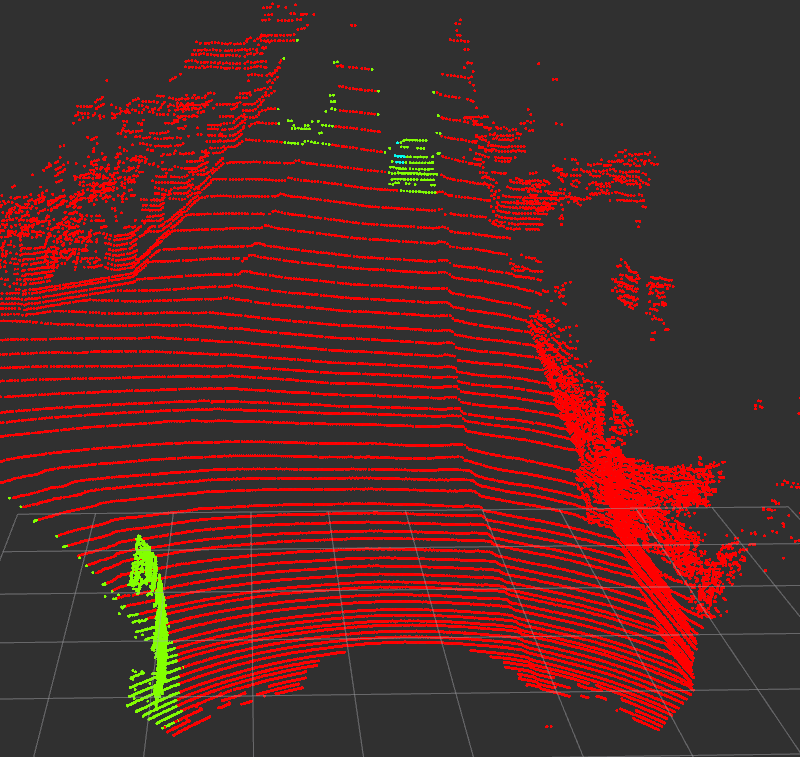}
\includegraphics[width=.245\textwidth, height =3.65cm]{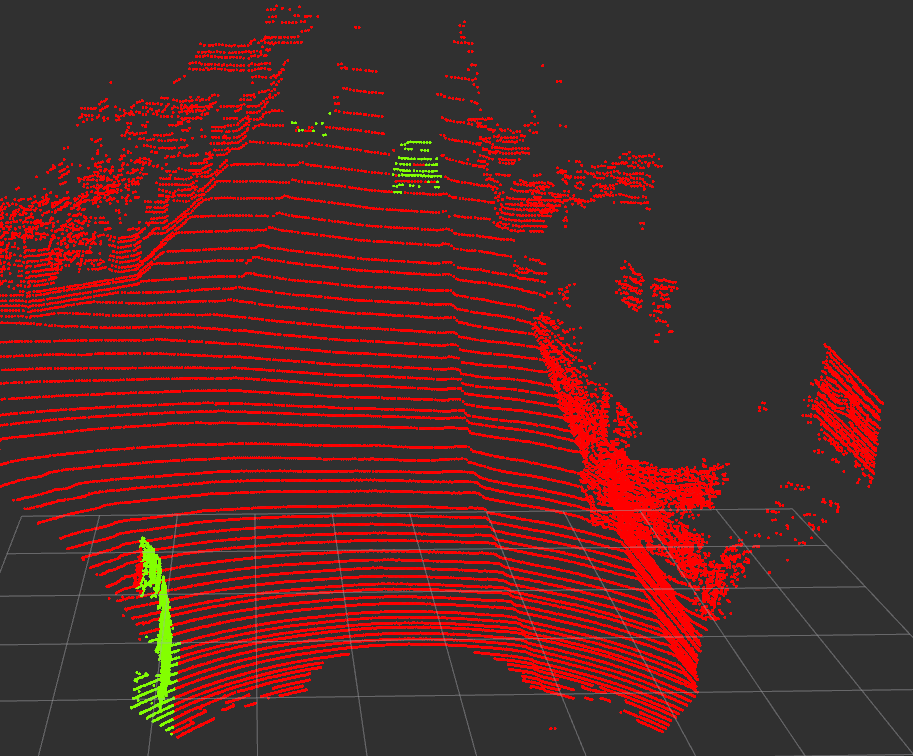}
\vspace{-3mm}

\includegraphics[width=.245\textwidth, height =5.15cm]{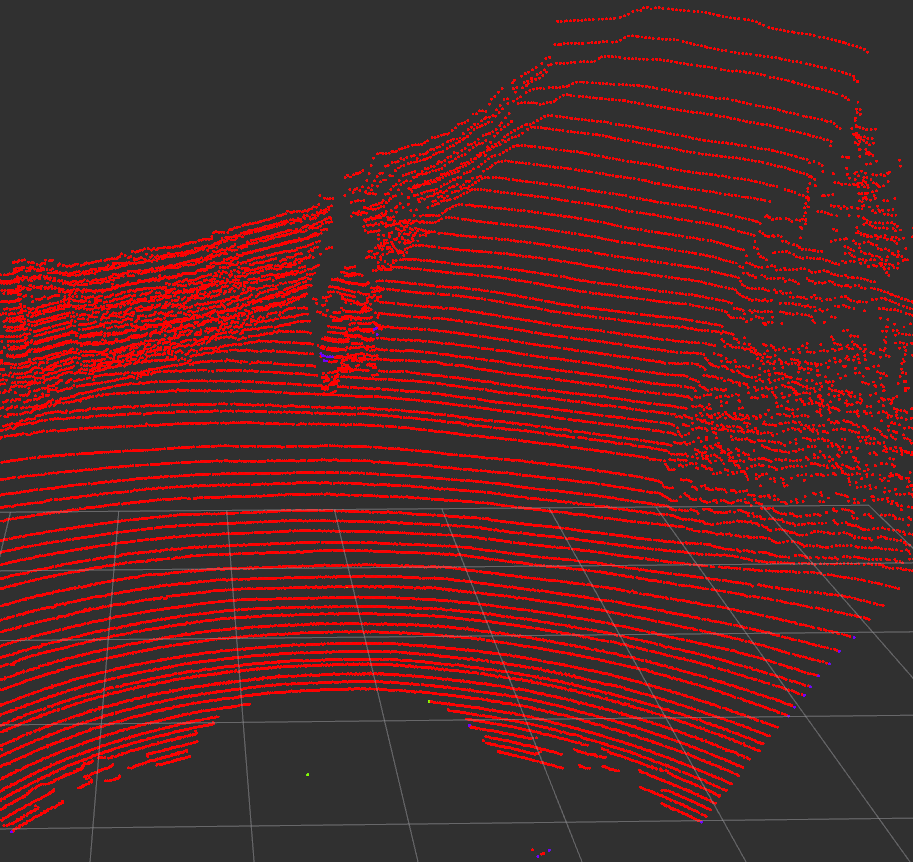}
\includegraphics[width=.245\textwidth, height =5.15cm]{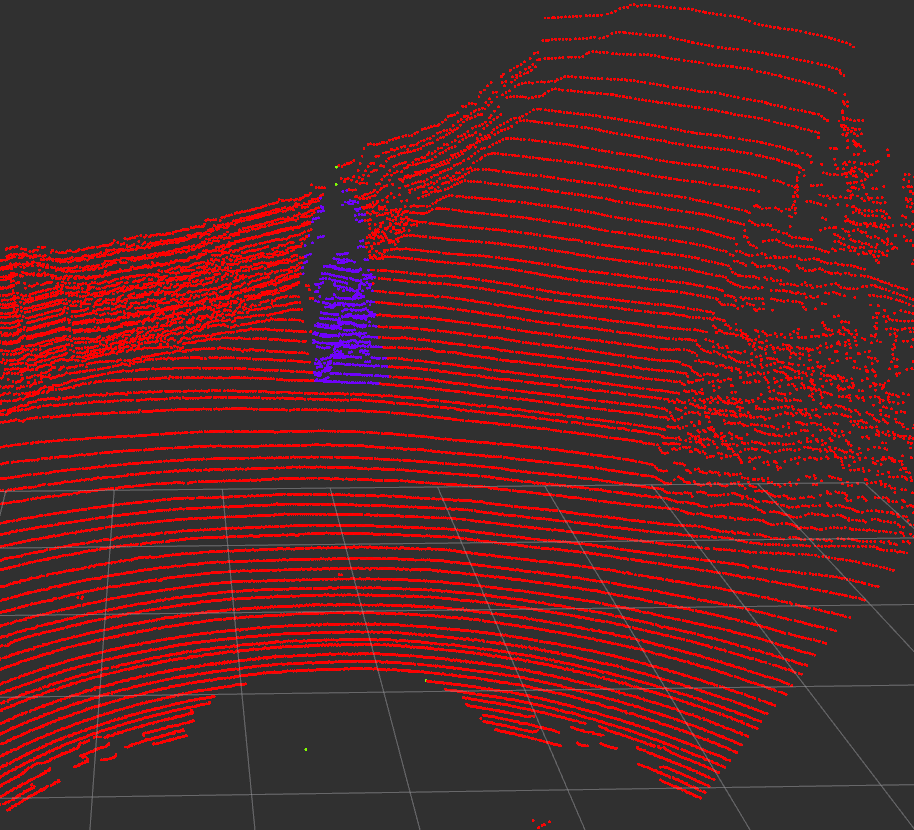}
\includegraphics[width=.245\textwidth, height =5.15cm]{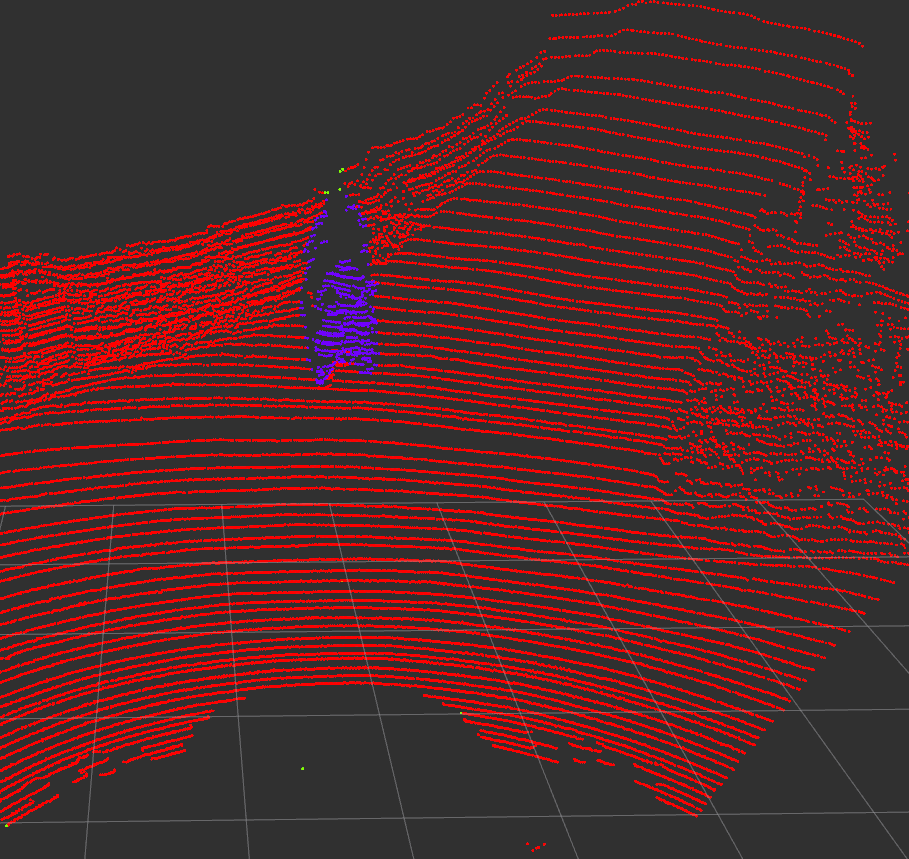}
\includegraphics[width=.245\textwidth, height =5.15cm]{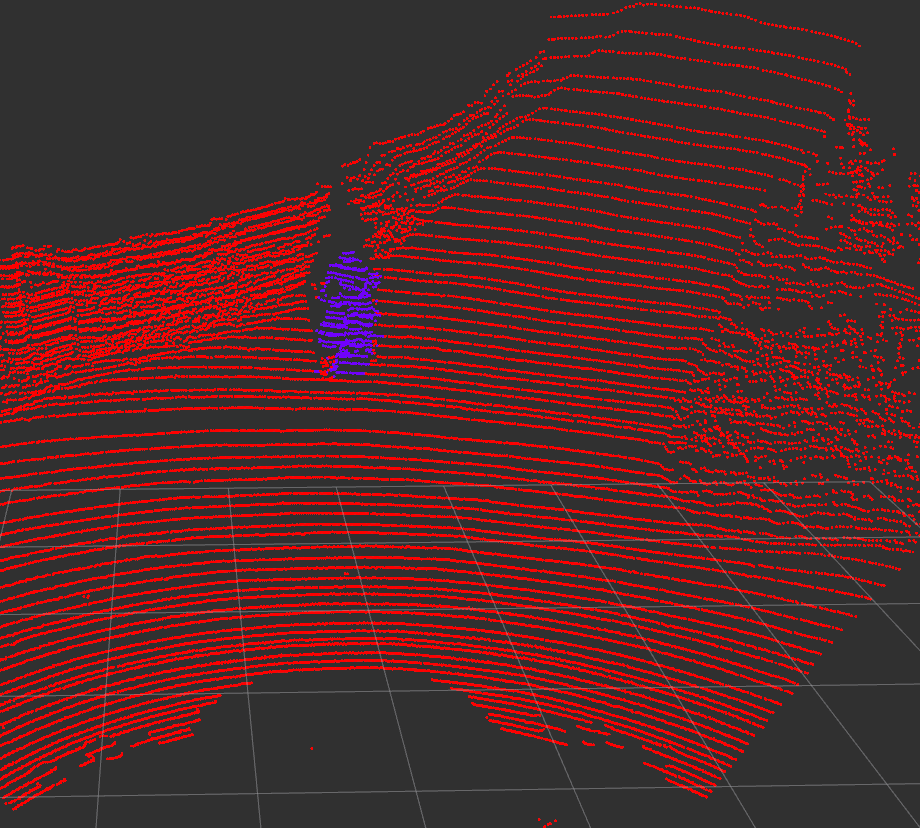}

\caption{Qualitative comparison of 3D semantic segmentation outputs using our approach on PointSeg architecture where red color represents None,green color represents Cars, violet represents Cyclists, and light-blue color represents Pedestrians. \textbf{Left:} No-Fusion output. \textbf{Middle-Left:} Early-Fusion output. \textbf{Middle-Right:} Mid-Fusion output. \textbf{Right:} Ground Truth.}
\label{fig:pointSeg_qualitative}
\end{figure*}

Fig. \ref{fig:pointSeg_qualitative}  shows qualitative comparison between the results obtained using PointSeg architecture, where there is improvement in cyclist and Cars.

% In Table \ref{tab:results}, we can notice that car class is better than pedestrian class and cyclist class, and due our analysis this is due to the data split didn't take in it's consideration the distribution of the three classes through the whole dataset, and however the PGM approach gives the DNN the ability to sense the geometric shapes of the surrounded environment, but the small volume of the two classes(pedestrians and cyclist) creates a new challenge for the DNN to be detected.

\end{document}